\documentclass[10pt,twocolumn,letterpaper]{article}

\usepackage{cvpr}              % To produce the CAMERA-READY version
\usepackage[dvipsnames]{xcolor}

\definecolor{cvprblue}{rgb}{0.21,0.49,0.74}
\usepackage[pagebackref,breaklinks,colorlinks,citecolor=cvprblue]{hyperref}
\usepackage{multirow}
\usepackage{multicol}
\usepackage{csquotes}
\def\paperID{7917} % *** Enter the Paper ID here
\def\confName{CVPR}
\def\confYear{2024}

\title{Retrieval-Augmented Embodied Agents}

\author{First Author\\
Institution1\\
Institution1 address\\
{\tt\small firstauthor@i1.org}
\and
Second Author\\
Institution2\\
First line of institution2 address\\
{\tt\small secondauthor@i2.org}
}

\author{
$\text{Yichen Zhu}$, $\text{Zhicai Ou}$, $\text{Xiaofeng Mou}$, $\text{Jian Tang}\thanks{Corresponding author}$
\\
$\text{Midea Group, AI Lab}$ \\
\tt\small \{zhuyc25, zhicai.ou, mouxf, jiantang22\}@midea.com}

\begin{document}
\maketitle
\begin{abstract}
Embodied agents operating in complex and uncertain environments face considerable challenges. While some advanced agents handle complex manipulation tasks with proficiency, their success often hinges on extensive training data to develop their capabilities. In contrast, humans typically rely on recalling past experiences and analogous situations to solve new problems. Aiming to emulate this human approach in robotics, we introduce the Retrieval-Augmented Embodied Agent (RAEA). This innovative system equips robots with a form of shared memory, significantly enhancing their performance. Our approach integrates a policy retriever, allowing robots to access relevant strategies from an external policy memory bank based on multi-modal inputs. Additionally, a policy generator is employed to assimilate these strategies into the learning process, enabling robots to formulate effective responses to tasks. Extensive testing of RAEA in both simulated and real-world scenarios demonstrates its superior performance over traditional methods, representing a major leap forward in robotic technology.

\end{abstract}

\section{Introduction}
The swift advancement of foundation models in areas like natural language processing and computer vision has sparked interest in the robotics community to create embodied agents capable of comprehending human instructions and responding aptly to their environment. Despite this enthusiasm, crafting agents that seamlessly interact with the physical world remains a formidable task. deep neural networks store knowledge—such as recognizing objects or interpreting commands—implicitly within their neural network parameters. This dependence on implicit knowledge storage demands a significant number of parameters and a wide range of training data. However, recent studies have shown that the scalability in terms of both training data and model size falls short~\cite{zitkovich2023rt2, bousmalis2023robocat} when compared to foundation models in other domains, such as Large Language Models. This insight has inspired the creation of embodied agents designed to learn efficiently with limited data and model sizes. To augment their capabilities, it's becoming increasingly important for these agents to access external repositories of physical knowledge, thereby expanding their capacity to understand and interact with the world.
\begin{figure}[t]
    % \begin{minipage}{1.0\linewidth}
    \centering
    \includegraphics[width=0.5\textwidth]{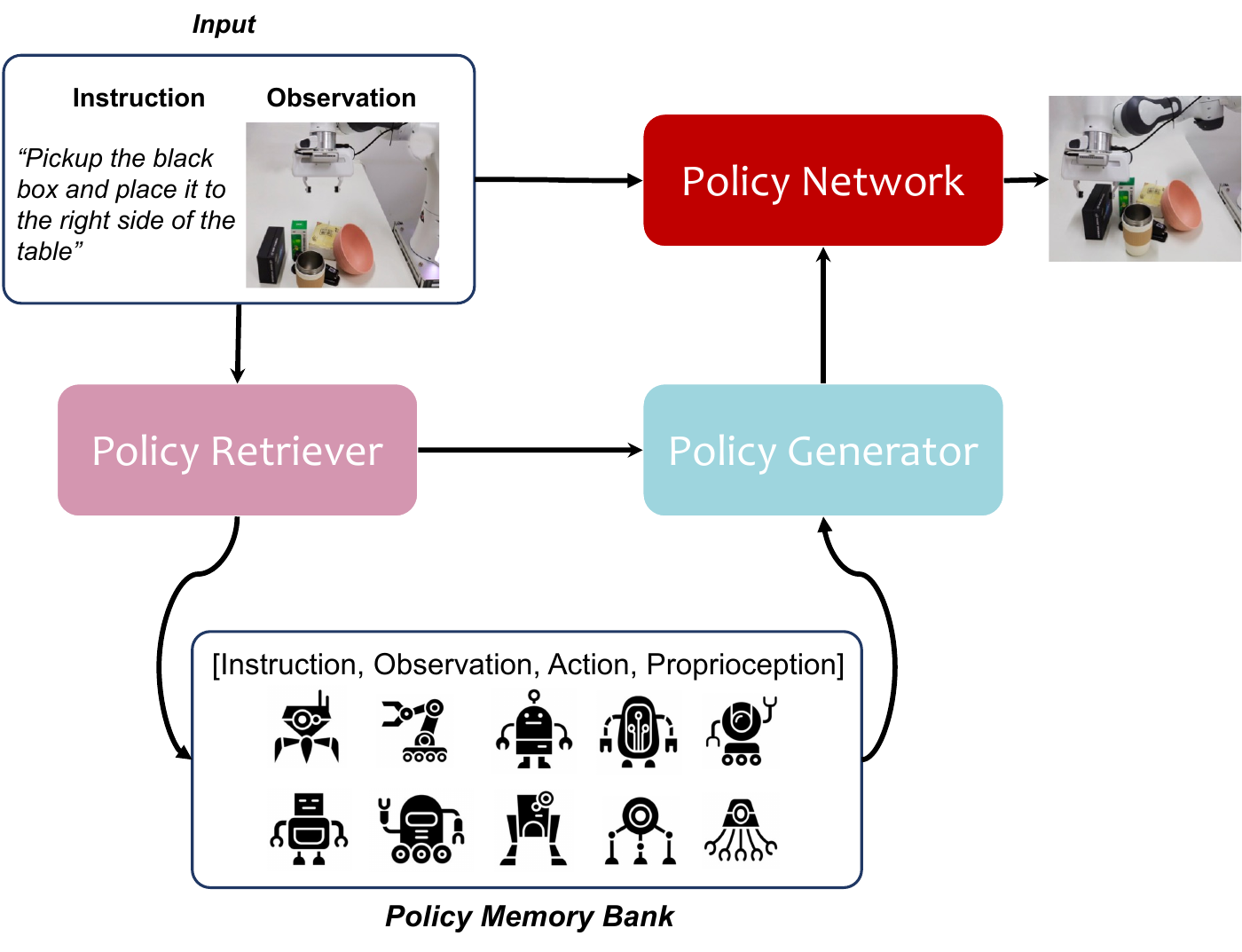}
    \caption{The overview of our retrieval-augmented embodied agents. We utilize a policy retriever to extract policies from a policy memory bank, which contains large-scale robotic data across multiple embodiments. Then, we use the policy generator to reference the retrieved policy and output actions for the current input.}\label{fig:exp_setup}
\end{figure}

The ability to tap into an external repository of behavioral memory mirrors the learning process observed in human infants, who often remember and mimic the actions of adults or animals when presented with analogous scenarios from their memory.  This ability is crucial for successfully navigating unknown environments and performing tasks that demand specific knowledge, like exploring unfamiliar rooms or handling new objects. Consequently, the question naturally arises: How can we harness the wealth of open-source, multi-embodiment data to enhance the precision of robots in manipulation tasks? This inquiry not only probes the potential of robotic learning but also seeks to bridge the gap between human cognitive processes and robotic applications.

\begin{figure*}[h]
    % \begin{minipage}{1.0\linewidth}
    \centering
    \includegraphics[width=\textwidth]{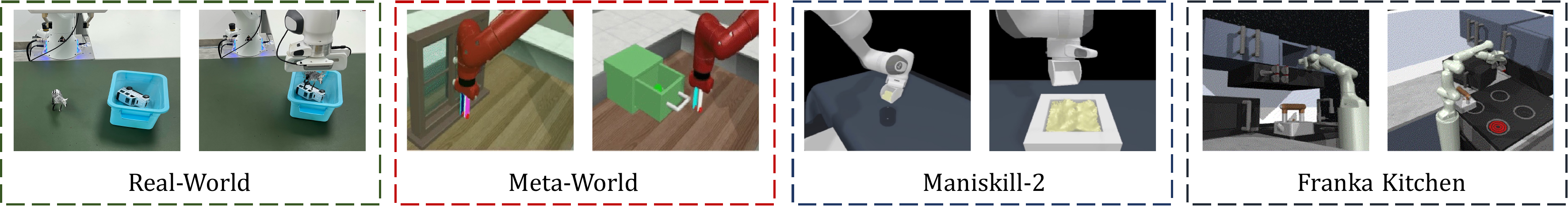}
    \caption{Examples of simulated and real-world environments that we used for evaluation.}\label{fig:four_data_visual}
\end{figure*}
In this paper, we present Retrieval-Augmented Embodied Agents (RAEA), which leverage an external policy memory bank containing analogous scenarios, whether in terms of instructions, observations, or a combination of both, related to the ongoing task. We outline the overview of our framework in Figure~\ref{fig:exp_setup}. The policies retrieved from this memory, along with other relevant data, provide a rich resource for both the training and testing phases. In our methodology, we make use of the recently open-sourced Open X-Embodiment~\cite{padalkar2023openx}, a large-scale repository of robotic datasets. Open X-Embodiment contains an extensive array of tasks, applications, embodiments, and diverse environmental settings from various research labs. This extensive dataset serves as the cornerstone for our external policy banks, enriching the knowledge base of RAEA. By tapping into this vast repository, RAEA can access a broader spectrum of robotic experiences, thereby improving its adaptability and effectiveness in various tasks.

To realize our objectives, we introduce two innovative modules: a policy retriever and a policy generator. The policy retriever is adept at handling multiple input modalities, categorized into two main types: instructions and observations. For controlling robots, it accepts text and audio as instructional inputs, and images, videos, and point clouds as observational inputs. It identifies policy candidates from the memory bank that align closely with the current input. Furthermore, we have developed the policy generator, which initially processes the information in the retrieved policies, including observation, instruction, action, and proprioceptive state. It then employs a cross-attention module to integrate knowledge from various retrieved policies into the main policy networks for action prediction. In this way, the policy generator leverages the retrieved policies as contextual examples, aiding the model in producing actionable responses based on the current input.

The efficacy of our proposed Retrieval-Augmented Embodied Agent (RAEA) is demonstrated through extensive testing over two simulation benchmarks and real-world datasets, as illustrated in Figure~\ref{fig:four_data_visual}. This approach not only showcases the versatility and practical ability of our framework.

In summary, our contributions are as follows:
\begin{itemize}
    \item We present Retriveal-Augmented Embodied Agents (RAEA) that utilized the wealth of knowledge from an external policy memory bank with multiple embodiment data to facilitate prediction for robotic action.
    \item Our framework features a policy retriever adept at processing various input modalities. Complementing this, we have crafted a policy generator that leverages retrieved scenarios to improve the model's ability to generalize across various situations.
    \item To validate the efficacy of our proposed methodology, we have conducted extensive evaluations in both real-world settings and two simulated environments. The results from these tests strongly affirm the effectiveness and practicality of our approach.
\end{itemize}
Overall, our work introduces a versatile and modular retrieval-augmentation framework for embodied agents. This provides a novel and insightful perspective on the design of robotic models, integrating advanced memory capabilities.

\section{Related Works}
\noindent
\textbf{Retrieval-Augmented Models.} A notable trend in Natural Language Processing (NLP) involves leveraging external memory to enhance the performance of language models. This approach retrieves documents relevant to the input text from an external database, allowing language models (generators) to use this retrieved information to make more informed predictions. Typically, the external memory consists of a collection of text passages or a structured knowledge base~\cite{yasunaga2022linkbert, yasunaga2021qa, xie2022unifiedskg}. Subsequent works extend the retrieval augmentation techniques to computer vision models and multi-modal models. The most representative works including Re-Imagen~\cite{chen2022reimagen} as a caption-to-image generator, MuRAG~\cite{chen2022murag} performs question-answering using retrieved images. RA-C3M~\cite{racm3} uses retrieval for either text or image generation.RA-CLIP~\cite{xie2023raclip} and REACT~\cite{liu2023ra} integrate the retrieval-augmentation technique for CLIP pretraining. Our approach, however, diverges from these prior works. While the aforementioned studies focus on enhancing language and vision models, our research is specifically geared towards robotics. We aim to search for policies that have been executed in scenarios similar to the current context, using them as in-context examples.
\\
\\
\noindent
\textbf{Models for Embodied Agents.} In the field of embodied agents~\cite{brohan2022rt1, nair2022learning, shao2021concept2robot, lynch2020language, wang2023d, huang2023voxposer, cui2022can, cui2022play, pathak2018zero, nasiriany2019planning, bahl2022human, bahl2023affordances, bertasius2016first, chen2021learning, sermanet2018time, sharma2019third, smith2019avid, shah2023mutex, jiang2022vima} and robotics~\cite{shapebias, dino, ebert2021bridge, hansen2022pre, oquab2023dinov2, parisi2022unsurprising, walke2023bridgedata, yuan2022pre}, foundation models have become a crucial research focus, revolutionizing the interaction between AI systems and physical environments. This body of work includes studies on representation pre-training and the application of language and vision-language models~\cite{hill2020human, jang2022bc, jiang2022vima, lynch2020language, nair2022learning, reed2022generalist, shridhar2023perceiver} as embodied agents. Our research contributes to this growing body of knowledge, presenting a supplementary retrieval-based approach designed to enhance policy learning in robotics. This approach integrates with existing foundation models, providing a novel perspective on how to augment the capabilities of embodied agents in diverse and dynamic settings.
\\
\\
\noindent
\textbf{Robotics Datasets.} The robotics learning community has developed a variety of open-source datasets that are instrumental in advancing robot learning. These tasks spanning grasping~\cite{kalashnikov2018qt, levine2018learning, depierre2018jacquard, mahler2017dex, eppner2021acronym, fang2020graspnet, brahmbhatt2019contactdb}, pushing interactions~\cite{shilane2004princeton, ebert2018visual, yu2016more},
sets of objects and models [75–85], and teleoperated demonstrations [8, 86–95]. Typically, these datasets are extensive and are often focused on specific robotic embodiments, exemplified by the Bridge Data~\cite{ebert2021bridge, walke2023bridgedatav2, kumar2022pre} and RH20T~\cite{fang2023rh20t}. RoboNet~\cite{dasari2019robonet} and Open X-Embodiment~\cite{padalkar2023openx} stand out as two large-scale datasets that incorporate multiple robotic embodiments. These datasets are frequently utilized for pretraining purposes~\cite{padalkar2023openx}, especially for the visual backbone in robotic models~\cite{burnspre, nair2022r3m, sensorpre}. In our research, we leverage the extensive cross-embodiment data as a foundational knowledge base. This enables us to retrieve relevant policies that facilitate training in the current environment, effectively utilizing the rich diversity of the datasets to enhance our model's adaptability and performance.

\section{Methodology}
\begin{figure*}[t]
    % \begin{minipage}{1.0\linewidth}
    \centering
    \includegraphics[width=0.9\textwidth]{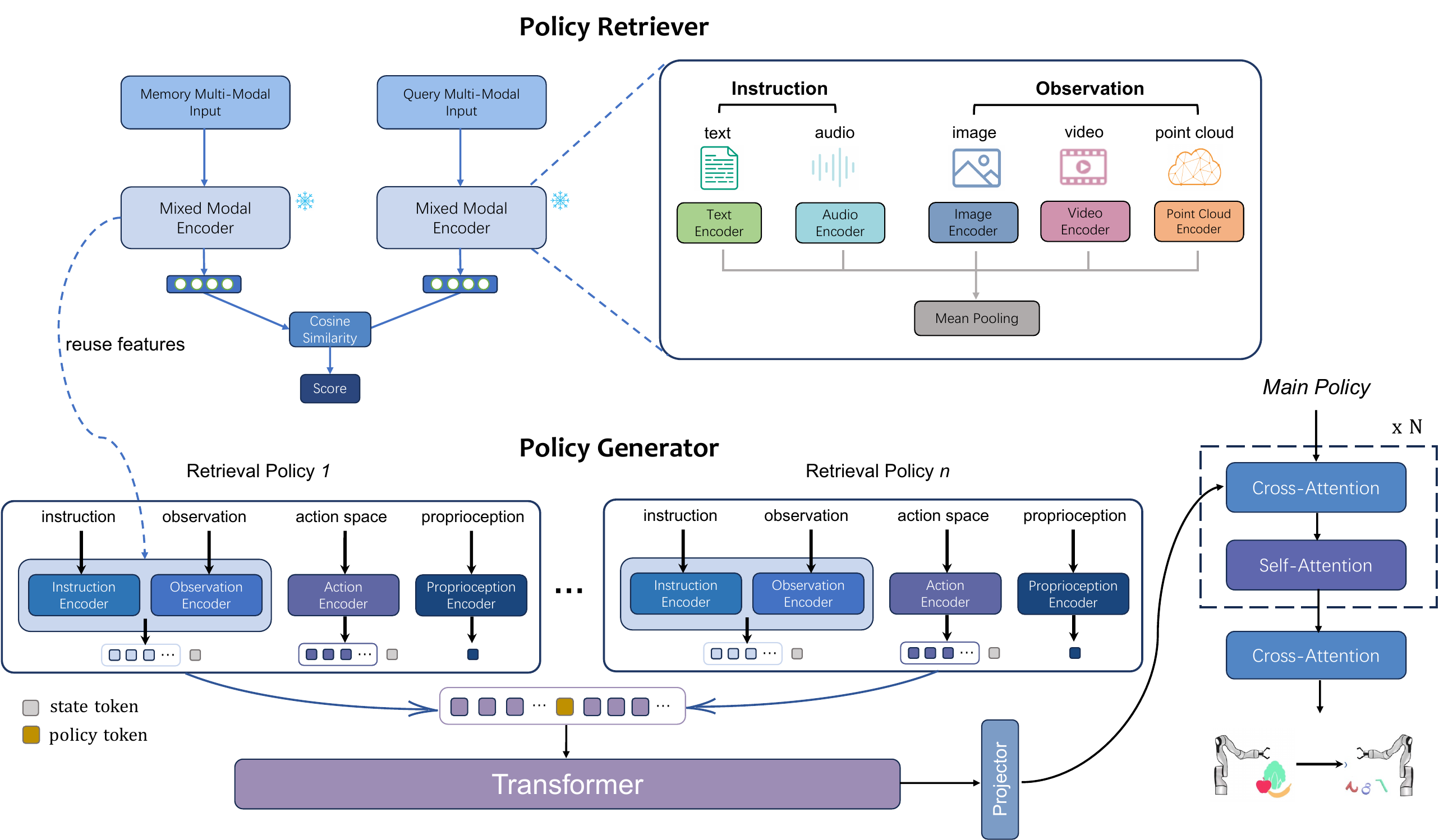}
    \caption{The framework of policy retriever (top) and policy generator (bottom) in our work. The policy retriever retrieves the relevant policy based on multi-modal input, and the policy generator processes a list of retrieved policies to help train in the current environment.}\label{fig:policy_re_gen}
\end{figure*}

We introduce Retrieval-Augmented Embodied Agents, capable of retrieving relevant scenarios and generating actions based on the current scene and accompanying instructions. As illustrated in Figure~\ref{fig:policy_re_gen}, when presented with an input, be it an observation or an instruction-observation pair, our system employs a retriever to fetch pertinent policies from an external memory bank. Significantly, these embodied agents are equipped to interact with humans through various modalities like text and audio, and they use diverse sensors to perceive their environment. In order to broaden the spectrum of applications for our approach, we have designed a multi-modal policy retriever (in \S 3.2), featuring a dense retriever with a mixed-modal encoder capable of encoding diverse modalities in various combinations. Additionally, we've constructed a policy generator (in \S 3.3)based on the Transformer architecture. This generator processes the retrieved policies individually and leverages cross-attention to incorporate the extracted information from the retrieved policies into the primary model branch. 

\subsection{Preliminaries}
\noindent
\textbf{Notations}. The framework consists of a policy retriever $R$ and a policy generator module $G$. The retrieval module $R$ takes an input sequence $r = \{i, o\}$ and searches the $r$ from an external policy memory bank. It returns a list of policy $m = \{i, o, a, p\}$, where $p$ is the policy, $i$ represents the instruction, $o$ denotes the observation, $a$ is the action, $p$ is the proprioception. The term proprioceptive robot state is used to describe a robot's intrinsic awareness of its own positioning and movement within a given space, which includes factors like joint angles, velocity, torque, and other physical statuses. The term actions refers to the specific tasks or operations executed by the robot. The policy generator $G$ then takes the input sequence $x$ and the retrieved policy $M = \{m_{1}, m_{2}, \cdots, m_{n}\}$ and returns the action $a$, where $a$ represent continuous actions that control the robots.

\subsection{Policy Retriever}
\noindent
\textbf{Overview.} A policy retriever $R$ takes a query $q$ (e.g., the instruction-observation pair x) with multi-modal information from the policy memory bank $M$, and returns a relevance score $r(q, m)$. We follow prior retrieval works~\cite{karpukhin2020dense}, in which the retriever $r$ is a bi-encoder architecture,
\begin{equation}
    r(q, m) = E_{Q}(q)^{T}E_{M}(m)
\end{equation}
Here, we employ two key encoders: $E_{Q}$, responsible for encoding queries, and $E_{M}$, which encodes the memory to yield dense vectors representing query and memory policies, respectively. Given that our input and memory consist of multi-modal documents, we leverage $E_{Q}$ and $E_{M}$ as mixed-modal encoders capable of handling various modalities. The architecture of mixed-modal encoders can be designed in multiple ways. In our specific context, we introduce a multi-modal retrieval approach that adeptly accommodates multiple modalities. When dealing with a multi-modal input for policy learning, we partition it into two distinct components: an instruction segment and an observation segment. The instruction segment typically contains human instructions in different formats, including text or audio, while the observation segment can encompass images, videos, or point cloud data. Each modality type is separately encoded using off-the-shelf, pre-trained multi-modality encoders, which we will discuss specifically in the next section. 
\\
\\
\noindent
\textbf{Multi-Modal Encoders.} The primary aim of multi-modal encoders is to equip embodied agents with the capability to handle a wide range of modalities, adapting to various scenarios. Specifically, we conceptualize the encoder as a mapping function, denoted by $P(\cdot)$. Typically, for each modality, a specialized model is needed to extract useful modality-specific representations. These are then projected onto a feature plane, ensuring uniformity in the shape of feature tensors across all modalities through projection layers. Following this, we average the vectors representing these modalities, normalizing their L2 norms to 1, thus generating a consolidated vector representation of the document. This encoding technique is uniformly applied to both $E_{Q}$ and $E_{M}$. The final step involves assessing the similarity of cross-modality features between the query and the memory.

There are a number of options~\cite{wu2023nextgpt} for the multi-modal model when it comes to handling different modalities. For example, CLIP~\cite{clip} or T5~\cite{raffel2020exploring} could be used for text and image processing. In our case, we utilize ImageBind~\cite{girdhar2023imagebind}, a high-performance encoder proficient across six modalities, for processing diverse input types. With the help of ImageBind, we are spared from managing many numbers of heterogeneous modal encoders. This mapping function $P(\cdot)$ maps all modalities to a unified latent embedding, greatly enhancing the efficiency of comparing feature similarity for retrieval purposes. For the retrieval process, we execute the Maximum Inner Product Search within the memory space, yielding a ranked list of candidates based on their relevance scores. From this list, we select the final $k$ policies for further analysis and processing.
\\
\\
\noindent
\textbf{Retrieval Strategy.}  We discuss three key factors in obtaining/sampling informative retrieved policies for the generator in practice. 
\\
\\
\noindent
\textbf{\textit{Relevance:}} The retrieval of policies must be closely aligned with the input sequence, covering either instructions, observations, or both. Without this alignment, the retrieved policies fail to provide meaningful contributions to the modeling of the primary input sequence. To ascertain the relevance of these policies, we employ a dense retriever score based on modality encoders. 
\\
\\
\noindent
\textbf{\textit{Input Modality:}} Our methodology divides the input into two distinct segments: instructions and observations, each tailored to process specific input modalities. The observation component is adept at recalling comparable scenes or objects, thereby enabling the model to comprehend scenarios not encountered in its training dataset. Concurrently, the instruction segment recapitulates actions previously executed during training. This flexible framework allows for either independent or combined usage of these segments, contingent on the computational resources and specific application contexts. This approach bears a resemblance to in-context learning, wherein the Large Language Model (LLM) is presented with scenarios that are similar, albeit not identical, to enhance response quality. Typically, we employ instruction-observation pairs at the onset of a frame, subsequently relying solely on the observation for the remainder of the frame until the next user interaction. It's noteworthy that this strategy can be integrated with recent advancements in Large Multi-Modal Models, facilitating real-time corrections of the robot's actions.

\noindent
\textbf{\textit{Diversity:}} We discovered that diversity in the policy memory bank is crucial for effective performance. Selecting the top-ranked actions based on relevance scores often leads to the inclusion of duplicate or very similar instruction and observation. This redundancy can detrimentally impact the performance of the generator. This challenge is particularly acute given that our action bank comprises fragments of videos. To optimize the efficacy of retrieval-based methods, it's essential to ensure a diverse range of in-context samples. These diverse samples are instrumental in aiding policy networks to effectively learn from demonstrations, both in training and testing stages. To address the issue of redundancy, our approach involves bypassing a candidate action if its relevance score is too closely aligned (e.g., exceeding 0.9) with the query or actions already retrieved. Additionally, to further enhance diversity, we propose a unique strategy of random token dropout from the query used in retrieval, approximately 70\% of tokens. This approach serves as a regularization mechanism during training and has been observed to significantly improve the generator's performance. We also include embodiment data of embodiment that are different from the robot that we used for evaluation. We observed that even with variations in embodiment, this diverse dataset still aids the policy network in its learning process.
\\
\\
\noindent
\textbf{Data Format of Retrieved Policy.} For every policy retriever, our framework covers a set of elements $m$ that either influence or result from control processes. Subsequently, we introduce our policy generator and demonstrate how it effectively utilizes this diverse spectrum of information for enhancing policy learning.
\begin{figure*}[t]
    % \begin{minipage}{1.0\linewidth}
    \centering
    \includegraphics[width=\textwidth]{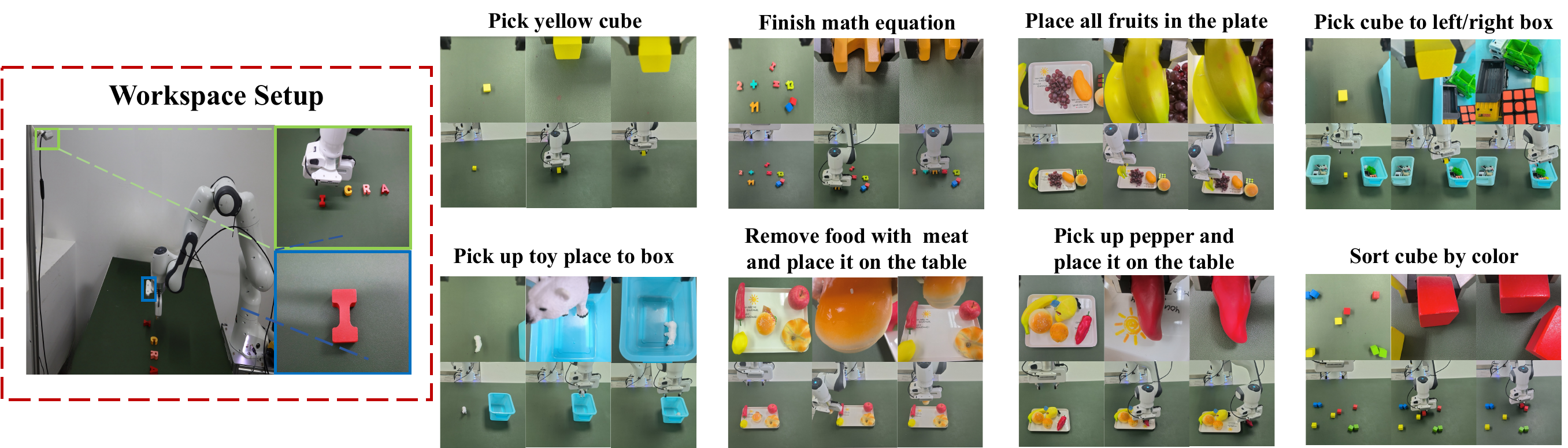}
    \caption{\textbf{Left:} The setup of our Franka real robot. \textbf{Right:} The example of some tasks that we collected.}\label{fig:realworld_data}
\end{figure*}
\begin{figure*}[t]
    % \begin{minipage}{1.0\linewidth}
    \centering
    \includegraphics[width=\textwidth]{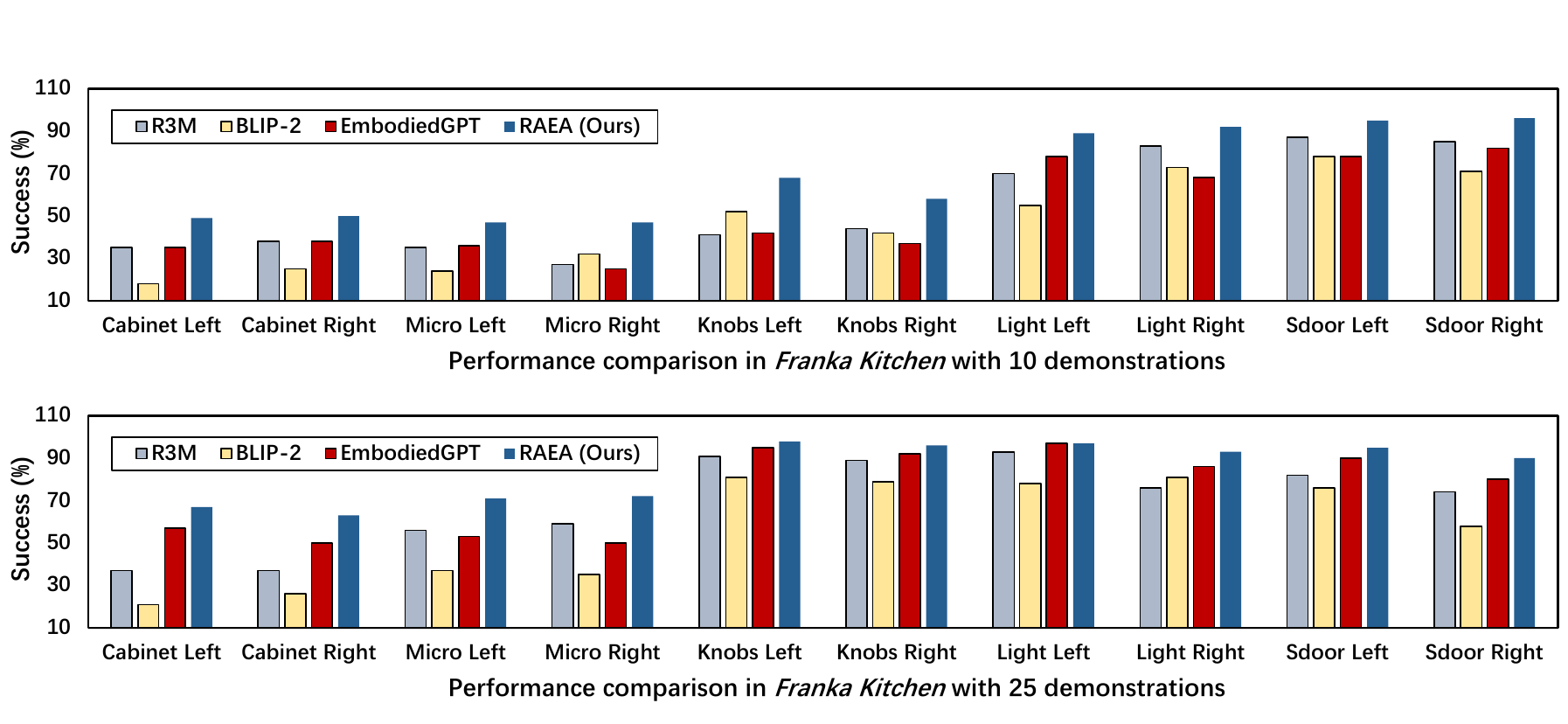}
    \caption{Performance of RAEA in Franka Kitchen with 10 or 25 demonstrations}\label{fig:franka}
\end{figure*}
\subsection{Policy Generator}
The policy generator is designed to effectively utilize the valuable information in the retrieved policy to facilitate the training of the policy for the current input. We reuse the feature representation of instruction and observation from the policy retrieval network (as in \S 3.2), thus avoiding redundant computations that constitute over 95\% of the total. For actions and proprioceptive states, we address the variability across different robots by setting a maximum limit for both, capped at nine. We employ an action encoder and a proprioception encoder – both comprising multi-layer perceptrons (MLPs) – to generate corresponding tokens. These tokens are then integrated with the instruction-observation tokens, ensuring a seamless and effective incorporation of the retrieved policy’s data into the current policy training framework. We add a state token between different states, i.e., a learnable token between action and proprioception tokens, to split the data of two states. We use absolute position embedding to ensure the tokens are in order.

Given a list of retrieved policies $M = (m_{1}, ..., m_{K})$, we concatenated these tokens based on the relevance score. We use absolute position embedding to maintain the order of tokenized representations. We concatenate their tokens according to their relevance scores. We employ a policy token to demarcate tokens from different policies. Once these tokens are combined, we utilize the Transformer architecture as the foundation for our retrieved policy processor. To effectively incorporate the retrieved policies $M$ into the generator, we leverage cross-attention mechanisms. This approach allows for the integration of pertinent information from the retrieved models into the main network, enhancing the overall efficacy of the system.

Particularly, give an input sequence from retrieved polices, denoted as $F^{r} \in \mathbb{R}^{h\times w\times c}$, and other input sequences from main input $F^{x} \in \mathbb{R}^{h\times w\times c}$, for simplicity, we assume two tensors have the same size. The $F^{r}$ is projected into a query (Q) and key (K), and $F^{x}$ is projected into value (V). Thus, we formulate our cross-attention (SC) as follows:
\begin{equation}
    Q_{i} = F^{r}W_{i}^{Q}
\end{equation}
\begin{equation}
    K_{i} = SC(F^{x}, r_{i})W_{i}^{K}, V_{i} = SC(F^{x}, r_{i})W_{i}^{V}, 
\end{equation}
\begin{equation}
 V_{i} = V_{i} + P(V_{i})
\end{equation} 
where $SC(\cdot, r_{i})$ is a MLP layer for aggregation in the $i^{th}$ head with the down-sampling rate of $r_{i}$, and $P(\cdot)$ is a depth-wise convolutional layer for projection. Finally, we calculated the attention tensor by:
\begin{equation}
    h_{i} = Softmax(\frac{Q_{i}K^{T}_{i}}{\sqrt{d_{h}}}V_{i})
\end{equation}
where $d_{h}$ is the dimension. The streamlined design of the cross-attention mechanism ensures that valuable representations from retrieved policies are effectively incorporated into the main network, enhancing policy learning for the current input. To optimize this process, we implement behavior cloning, using mean squared loss as our primary optimization objective.

\section{Experiments}
We evaluate models using multiple simulated benchmarks, including Franka Kitchen~\cite{fu2020d4rl}, MetaWorld~\cite{yu2020metaworld}, and Maniskill-2~\cite{gu2023maniskill2} and real-world environment. We show that our retrieval-augmented embodied agents significantly improve the generalization ability. Notably, we have taken precautions to ensure that the policy memory bank does not contain any data from the datasets used in our training and testing phases, thereby eliminating the possibility of bias or cheating in the test set.

\subsection{Simulation Experiments.}
We conduct our simulation on three benchmarks, Franka Kitchen~\cite{fu2020d4rl}, Meta-World~\cite{yu2020metaworld}, and Maniskill-2~\cite{gu2023maniskill2}. A brief summary of these benchmarks can be found in the Appendix.
\begin{figure*}[t]
    % \begin{minipage}{1.0\linewidth}
    \centering
    \includegraphics[width=1.0\textwidth]{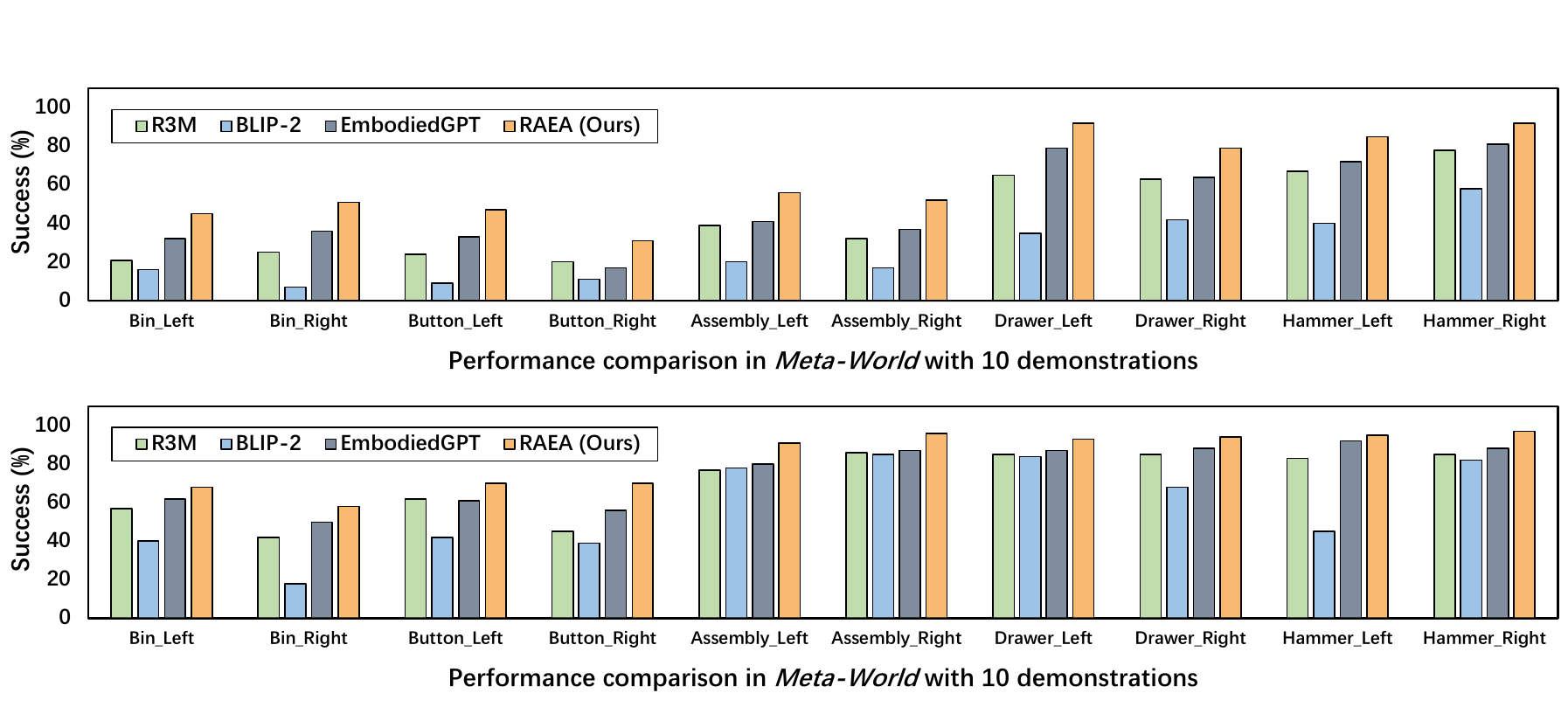}
    \caption{Performance of RAEA in Meta-World with 10 or 25 demonstrations}\label{fig:metaworld}
\end{figure*}
\\
\\
\noindent
\textbf{Evaluation}: We assess our approach through 30 rollouts derived from the behavior cloning (BC) learned policy. Our primary metric for evaluation is the mean success rate of the final policy. Additionally, when presenting metrics for the task suite, we calculate the average mean success rate across various camera configurations.
\\
\\
\noindent
\textbf{Experimental results on Franka Kitchen \& Meta-World.}  In our experiments, we conducted a comparative analysis of our model, RAEA, against two established state-of-the-art methods: R3M~\cite{nair2022r3m}, popular in Franka Kitchen applications, and BLIP-2~\cite{li2023blip}, a leading vision-language model, and Embodied-GPT~\cite{mu2023embodiedgpt}, a vision-language model designed for robotics. We trained our policy network using a few-shot learning approach, employing datasets comprising either ten or twenty-five demonstrations. The performance of these models was evaluated through 100 randomized trials across five distinct tasks in each benchmark. These evaluations were executed under two different settings: each involved five separate runs and was conducted from two unique camera perspectives, relying solely on visual observations. The results, depicted in Figures~\ref{fig:franka} and ~\ref{fig:metaworld}, for the Franka Kitchen and Meta-World benchmarks respectively, unequivocally demonstrate that RAEA surpasses the baseline methods in effectiveness. This superiority is particularly noticeable in low-data scenarios, such as those with only ten demonstrations, further underscoring RAEA's robustness and efficiency in environments with limited data availability.
\begin{table*}[t]
\renewcommand{\arraystretch}{1.3}
\caption{Experiments on Manisill-2 over six rigid body and soft body tasks. Our method consistently outperforms Baseline in all environments. All metrics are reported in percentage $(\%)$ with the best ones bolded.}
\centering
\resizebox{0.65\textwidth}{!}{\begin{tabular}{l|cccccc}
\toprule
Methods & PickCube & StackCube & PickSingleYCB & Fill & Hang & Excavate  \\
\midrule
\multicolumn{7}{c}{Observation Modality: Image} \\
\midrule
ResNet152~\cite{he2016deep} & 40.1 & 86.3& 22.1 & 52.4 & 82.6& 12.0  \\
Swin-Base~\cite{liu2021swin}  & 41.3 & 83.5&  28.5& 49.0 & 82.7& 14.8\\
\textbf{RAEA} & \textbf{56.7} &\textbf{93.6} &\textbf{40.2} & \textbf{63.8}&\textbf{87.1} &\textbf{22.4} \\
\midrule
\multicolumn{7}{c}{Observation Modality: Point Cloud + Image} \\
\midrule
ResNet152 + PointNet~\cite{he2016deep, qi2017pointnet} &  56.6 & 90.7& 28.4 & 46.9 & 85.7& 18.4 \\
Swin-Base + PointNet~\cite{liu2021swin, qi2017pointnet} & 44.0 & 90.1& 30.6 & 45.5 & 86.4& 17.0\\
\textbf{RAEA} & \textbf{62.7} &\textbf{91.0} &\textbf{43.8} & \textbf{71.7}&\textbf{88.1} &\textbf{24.0} \\
\bottomrule
\end{tabular}}
\label{table:maniskill}
\end{table*}

\\
\\
\noindent
\textbf{Experimental results on Maniskill-2.} We evaluated our model's performance in two distinct experimental settings: one using solely image-based observations and the other combining images with point cloud data. For comparison, we benchmarked our model against well-established methods, specifically ResNet152~\cite{he2016deep} and Swin Transformer Base~\cite{liu2021swin}, both pre-trained on the ImageNet dataset. In the experiments that involved both images and point clouds, the feature representation of point cloud data for the baseline models was initially processed using PointNet~\cite{qi2017pointnet}. This representation was subsequently integrated with the image branch, following the procedure described in Maniskill-2. In both experimental scenarios, our method consistently surpassed the performance of the two baseline models, as illustrated in Table~\ref{table:maniskill}. Remarkably, in the setting that utilized dual-modal observations, our RAEA model demonstrated a significantly higher average success rate. This highlights the robust generalizability and effectiveness of our approach.

\begin{table}[t]
\renewcommand{\arraystretch}{1.3}
\caption{Ablation study on the effect of using different modalities for real-world environments.}
\centering
\resizebox{0.5\textwidth}{!}{\begin{tabular}{l|l|l|c}
\toprule
Method & Instruction & Observation& Success Rate\\ 
\midrule
\midrule
\multirow{7}{*}{RAEA} & Text & Image & 54\\
 & Text+Audio & Image & 56 \\
 & Text+Audio & Image + Point Cloud & 65\\
 & Image & Image& 58 \\
 & Image+Video & Image & 63\\
 & Image+Video & Image + Video & 64\\
 & Image+Video+Text & Image & 69 \\

\bottomrule
\end{tabular}}
\label{table:modalities_exp}
\end{table}

\subsection{Real-Robot Experiments}
\noindent
\textbf{Datasets.} Our collected dataset comprises $n = 40$ tasks. These tasks vary from straightforward pick-and-place actions, such as \enquote{pick up the yellow cube} to more complex contact-rich tasks like \enquote{open the drawer and put the pen inside,} as well as tasks such as \enquote{sort the cube with the same color.} There are 70 objects in the experiments. Each task is exemplified through 30 human-collected trajectories. Further, every task is annotated with 5 distinct instructions. Figure~\ref{fig:realworld_data} demonstrate some example and workspace setup for our real-world experiments.
\\
\\
\noindent
\textbf{Implementation details.} We use an AdamW optimizer, starting with an initial learning rate of 3e-5, and implement a weight decay of 1e-6. Our learning rate scheduler is designed to linearly decay, incorporating a warm-up phase that spans the initial 2\% of the total training duration. Additionally, we apply gradient clipping set at a value of 1.0 to maintain stability during training. To assess the efficacy of our approach, all experiments are conducted over 10 trials, from which we calculate the mean success rate.
\\
\\
\noindent
Additionally, we perform ablation studies to delve into various questions related to our model's performance and capabilities.

\begin{table}[t]
\renewcommand{\arraystretch}{1.3}
\caption{Ablation study on the effect of status information, i.e., action \& proprioceptive state, in real-world data. The experiments are conducted based on text-image pairs.}
\centering
\resizebox{0.4\textwidth}{!}{\begin{tabular}{l|l|c}
\toprule
Tasks & Status & Success Rate\\ 
\midrule
\multirow{3}{*}{RAEA} & All &  54 \\
 & w/o Proprioception & 39\\
 & w/o Action \& Proprioception & 36\\
\bottomrule
\end{tabular}}
\label{table:pa_ablation}
\end{table}

\begin{table}[t]
\renewcommand{\arraystretch}{1.3}
\caption{Ablation study on the data for policy memory bank using Franka-only or all embodiments.}
\centering
\resizebox{0.3\textwidth}{!}{\begin{tabular}{l|l|c}
\toprule
Tasks & Embodiments & Success Rate\\ 
\midrule
\multirow{2}{*}{RAEA} & All & 54 \\
 &Franka-Only & 48\\
\bottomrule
\end{tabular}}
\label{table:embod_exp}
\end{table}

\noindent
\textit{1. Does Utilizing Multiple Modalities Improve Generalizability?} Our embodied agents are equipped to support multi-modal inputs. This section examines the benefits of integrating multiple modalities. Table~\ref{table:modalities_exp} demonstrates that employing more available modalities can stably yield better performance. Notice that using the combination of language and visual as instruction significantly enhance the generalizability of the model, i.e., increase the success rate from 63 to 69. Also, adopting 3D information, such as point cloud, can be useful, which improves the success rate from 56 to 65. 
\\
\\
\noindent
\textit{2. Does Including More Status Information in Retrieved Policies Enhance Policy Learning?} While instruction and observation are fundamental components, our research also delves into the impact of incorporating proprioception and action data. As illustrated in Table~\ref{table:pa_ablation}, there is a discernible decrease in success rate when proprioception and action are omitted from the retrieved policy, dropping from 54 to 36. This finding highlights the critical role these elements play in augmenting the efficacy of our learning approach. %It also indicated that it is insufficient to simply adopt multi-modal input as the retrieved document for policy learning, the information of robots along with their states are crucial to learning.
\\
\\
\noindent
\textit{3. Is Retrieval Across Different Embodiments Beneficial?} In our experiments, we utilized the Open X-Embodiment as our primary policy memory bank, a dataset featuring a variety of embodiments. Our evaluation, depicted in Table~\ref{table:embod_exp}, focuses on the performance implications of using data exclusively from Franka robots compared to a multi-embodiment dataset. We observed a slight decline in performance with the Franka-only data. This could be attributed to the richer diversity of environments and commands present in the broader dataset.

\section{Conclusion}
Training with pre-defined datasets often leads to a limited scope of ability and knowledge acquisition. In this paper, we introduce Retrieval-Augmented Embodied Agent (RAEA), a novel framework that enhances an embodied agent through a policy retriever and generation process. The primary objective of this retrieval process is to efficiently and effectively harness valuable insights from a comprehensive dataset of experiences, thereby aiding the agent in achieving its goals more proficiently. Through multiple ablation studies, we have underscored the significance of the various components within RAEA. Overall, our RAEA methodology presents an innovative and practical approach to leveraging collective knowledge from diverse datasets of different embodiments.

\newpage
{
    \small
    \bibliographystyle{ieeenat_fullname}
    \bibliography{main}
}


\end{document}